# Ancient Wisdom, Modern Tools: Exploring Retrieval-Augmented LLMs for Ancient Indian Philosophy

**Priyanka Mandikal**
Department of Computer Science, UT Austin
mandikal@utexas.edu

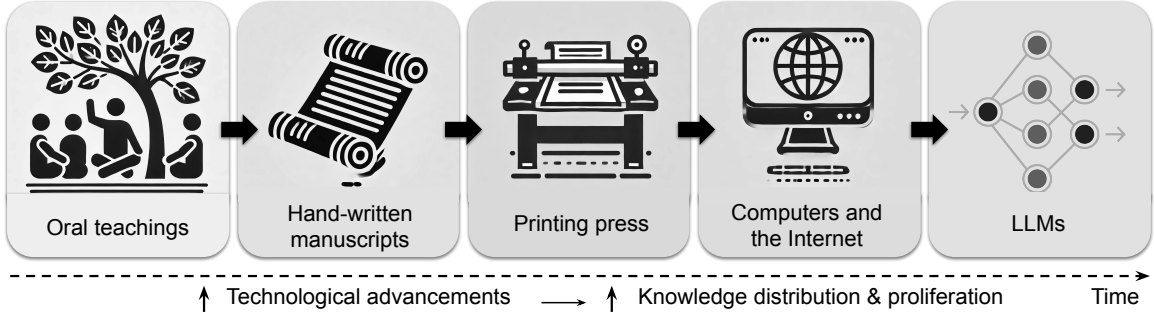

Figure 1: **The dissemination of knowledge through the ages.** Over time, methods of storing and transmitting knowledge have evolved from oral teachings to computers and the internet, significantly increasing the distribution and proliferation of human knowledge. The emerging LLM technology represents a new paradigm shift in this process.

## Abstract

LLMs have revolutionized the landscape of information retrieval and knowledge dissemination. However, their application in specialized areas is often hindered by factual inaccuracies and hallucinations, especially in long-tail knowledge distributions. We explore the potential of retrieval-augmented generation (RAG) models for long-form question answering (LFQA) in a specialized knowledge domain. We present VedantaNY-10M, a dataset curated from extensive public discourses on the ancient Indian philosophy of Advaita Vedanta. We develop and benchmark a RAG model against a standard, non-RAG LLM, focusing on transcription, retrieval, and generation performance. Human evaluations by computational linguists and domain experts show that the RAG model significantly outperforms the standard model in producing factual and comprehensive responses having fewer hallucinations. In addition, a keyword-based hybrid retriever that emphasizes unique low-frequency terms further improves results. Our study provides insights into effectively integrating modern large language models with ancient knowledge systems.

ॐ केनेषितं पतति प्रेषितं मनः ॥ 1.1 ॥
यन्मनसा न मनुते येनाहुर्मनो मतम् । तदेव ब्रह्म त्वं विद्धि नेदं यदिदमुपासते ॥ 1.6 ॥

"By whom willed and directed, does the mind alight upon its objects?"
"What one cannot comprehend with the mind, but by which they say the mind comprehends, know that alone to be Brahman, not this which people worship here."

– Kena Upanishad, >3000 B.C.E.

## 1 Introduction

Answer-seeking has been at the heart of human civilization. Humans have climbed mountains and crossed oceans in search of answers to the greatest questions concerning their own existence. Over time, ancient wisdom has travelled from the silent solitude of mountain caves and forest hermitages into the busy cities and plains of the world. Technology has played a major role in this transmission, significantly increasing the distribution and proliferation of human knowledge. In recent times, large language models (LLMs) trained on large swathes of the internet have emerged as de-facto question-answering machines. Recent studies on the societal impact of LLMs (Malhotra, 2021; Yiu et al., 2023) highlight their growing significance as cultural technologies. Analogous to earlier technologies like writing, print, and the internet, the power of LLMs can be harnessed meaningfully to preserve and disseminate human knowledge (Fig. 1).

Generic LLMs have proven to be highly effective for broad knowledge domains. However, they often struggle in niche and less popular areas, encountering issues such as factual inaccuracies and hallucinations in long-tail knowledge distributions (Kandpal et al., 2023; Mallen et al., 2023). Moreover, their inability to verify responses against authentic sources is particularly problematic in these domains, where LLMs can generate highly inaccurate answers with unwarranted confidence (Kandpal et al., 2023; Menick et al., 2022). In response to these limitations, there has been growing interest in retrieval-augmented generation (RAG) models (Karpukhin et al., 2020; Lewis et al., 2020b; Izacard et al., 2022; Ram et al., 2023). These models integrate external datastores to retrieve relevant knowledge and incorporate it into LLMs, demonstrating higher factual accuracy and reduced hallucinations compared to conventional LLMs (Shuster et al., 2021; Borgeaud et al., 2022). Updating these external datastores with new information is also more efficient and cost-effective than retraining LLMs. In this vein, we argue that RAG models show immense potential for enhancing study in unconventional, niche knowledge domains that are often underrepresented in pre-training data. Their ability to provide verified, authentic sources when answering questions is particularly advantageous for end-users.

In this work, we develop and evaluate a RAG-based language model specialized in the ancient Indian philosophy of Advaita Vedanta (Upanishads, >3000 B.C.E.; Bhagavad Gita, 3000 B.C.E.; Shankaracharya, 700 C.E.). To ensure that the LLM has not been previously exposed to the source material, we construct VedantaNY-10M, a custom philosophy dataset comprising transcripts of over 750 hours of public discourses on YouTube from Vedanta Society of New York. We evaluate standard non-RAG and RAG models on this domain and find that RAG models perform significantly better. However, they still encounter issues such as irrelevant retrievals, sub-optimal retrieval passage length, and retrieval-induced hallucinations. In early attempts to mitigate some of these issues, we find that traditional sparse retrievers have a unique advantage over dense retrievers in niche domains having specific terminology—Sanskrit terms in our case. Consequently, we propose a keyword-based hybrid retriever that effectively combines sparse and dense embeddings to upsample low-frequency or domain-specific terms.

We conduct an extensive evaluation comprising both automatic metrics and human evaluation by computational linguists and domain experts. The models are evaluated along three dimensions: transcription, retrieval, and generation. Our findings are twofold. First, RAG LLMs significantly outperform standard non-RAG LLMs along all axes, offering more factual, comprehensive, and specific responses while minimizing hallucinations, with an 81% preference rate. Second, the keyword-based hybrid RAG model further outperforms the standard deep-embedding based RAG model in both automatic and human evaluations. Our study also includes detailed long-form responses from the evaluators, with domain experts specifically indicating the likelihood of using such LLMs to supplement their daily studies. Our work contributes to the broader understanding of how emerging technologies can continue the legacy of knowledge preservation and dissemination in the digital age.

## 2 Related Work

**Language models for ancient texts** Sommerschield et al. (2023) recently conducted a thorough survey of machine learning techniques applied to the study and restoration of ancient texts. Spanning digitization (Narang et al., 2019; Moustafa et al., 2022), restoration (Assael et al., 2022), attribution (Bogacz and Mara, 2020; Paparigopoulou et al., 2022) and representation learning (Bamman and Burns, 2020), a wide range of use cases have benefitted from the application of machine learning to study ancient texts. Recently, Lugli et al. (2022) released a digital corpus of romanized Buddhist Sanskrit texts, training and evaluating embedding models such as BERT and GPT-2. However, the use of LLMs as a question-answering tool to enhance understanding of ancient esoteric knowledge systems has not yet been systematically studied. To the best of our knowledge, ours is the first work that studies the effects of RAG models in the niche knowledge domain of ancient Indian philosophy.

**Retrieval-Augmented LMs.** In current LLM research, retrieval augmented generation models (RAGs) are gaining popularity (Izacard et al., 2022; Ram et al., 2023; Khandelwal et al., 2020; Borgeaud et al., 2022; Menick et al., 2022). A key area of development in RAGs has been their architecture. Early approaches involved finetuning the language model on open-domain question-answering before deployment. MLM approaches such as REALM (Guu et al., 2020) introduced a two-stage process combining retrieval and reading, while DPR (Karpukhin et al., 2020) focused on pipeline training for question answering. RAG (Lewis et al., 2020b) used a generative approach with no explicit language modeling. Very recently, in-context RALM (Ram et al., 2023) showed that retrieved passages can be used to augment the input to the LLM in-context without any fine-tuning like prior work. In this work, we adopt the in-context retrieval augmented methodology similar to (Ram et al., 2023), where neither the retriever nor the generator is fine-tuned. This also enables us to use any combination of retrieval and generation models that best suits our application.

**Applications of RAGs.** The applications of RAGs are diverse and evolving. ATLAS (Izacard et al., 2022) and GopherCite (Menick et al., 2022) have shown how fine-tuning and reinforcement learning from human feedback can enhance RAGs' ability to generate verifiable answers from reliable sources. Prompting techniques have also seen innovation. kNNPrompt (Shi et al., 2022) extended kNN-LM for zero or few-shot classification tasks, and retrieval in-context approaches (Ram et al., 2023; Shi et al., 2023) have proven effective in utilizing retrieval at the input stage. Retrieval-LMs have been shown to be particularly valuable for handling long-tail or less frequent entities (Kandpal et al., 2023; Mallen et al., 2023), updating knowledge (Izacard et al., 2022), improving parameter efficiency (Izacard et al., 2022; Mallen et al., 2023), and enhancing verifiability (Bohnet et al., 2022), making them increasingly relevant in a wide range of applications. In our work, we examine the application of RAGs for long-tail knowledge, conducting an extensive study on a niche knowledge domain of ancient Indian philosophy.

**Evaluation of LFQA** The field of long-form question answering (LFQA) is an emerging area of active research (Krishna et al., 2021; Nakano et al., 2021; Xu et al., 2023). Recently, Xu et al. (2023) conducted a thorough examination of various LFQA metrics, encompassing both human and automatic evaluation methods, and found that existing automatic metrics don't always align with human preferences. On their suggestion, we place special emphasis on conducting an extensive human evaluation utilizing the expertise of experienced computational linguists and domain experts.

Figure 2: **Sanskrit terms in VedantaNY-10M.** Frequently occurring Sanskrit terms in the corpus.

## 3 The VedantaNY-10M Dataset

We first describe our niche domain dataset creation process. The custom dataset for our study needs to satisfy the following requirements: (**1**) **Niche:** Must be a specialized niche knowledge domain within the LLM's long-tail distribution. (**2**) **Novel**: The LLM must not have previously encountered the source material. (**3**) **Authentic:** The dataset should be authentic and representative of the knowledge domain. (**4**) **Domain experts:** should be available to evaluate the model's effectiveness and utility.

**Knowledge domain.** To satisfy the first requirement, we choose our domain to be the niche knowledge system of Advaita Vedanta, a 1300-year-old Indian school of philosophy (Shankaracharya, 700 C.E.) based on the Upanishads (>3000 B.C.E.), Bhagavad Gita (3000 B.C.E.) and Brahmasutras (3000 B.C.E.)[1]. It is a contemplative knowledge tradition that employs a host of diverse tools and techniques including analytical reasoning, logic, linguistic paradoxes, metaphors and analogies to enable the seeker to enquire into their real nature. Although a niche domain, this knowledge system has been continuously studied and rigorously developed over millenia, offering a rich and structured niche for the purposes of our study. Being a living tradition, it offers the additional advantage of providing experienced domain experts to evaluate the language models in this work.

**Composition of the dataset.** Considering the outlined criteria, we introduce VedantaNY-10M, a curated philosophy dataset of public discourses.

---

[1]Currently there exists no consensus on accurately dating these ancient scriptures. The Upanishads (which are a part of the Vedas) have been passed on orally for millennia and are traditionally not given a historic date. However, they seem to have been compiled and systematically organized sometime around 3000 B.C.E. by Vyasa. Likewise, the time period of Adi Shankaracharya also varies and he is usually placed between 450 B.C.E to 700 C.E.

To maintain authenticity while ensuring that the LLM hasn't previously been exposed to the source material, we curate our dataset from a collection of YouTube videos on Advaita Vedanta, sourced from the Vedanta Society of New York. It contains 10M tokens and encompasses over 750 hours of philosophical discourses by Swami Sarvapriyananda, a learned monk of the Ramakrishna Order. These discourses provide a rich and comprehensive exposition of the principles of Advaita Vedanta, making them an invaluable resource for our research.

**Languages and scripts.** The dataset primarily features content in English, accounting for approximately 97% of the total material. Sanskrit, the classical language of Indian philosophical literature, constitutes around 3% of the dataset. The Sanskrit terms are transliterated into the Roman script. To accommodate the linguistic diversity and the specific needs of the study, the dataset includes words in both English and Sanskrit, without substituting the Sanskrit terms with any English translations. Translating ancient Sanskrit technical terms having considerably nuanced definitions into English is a non-trivial problem (Malhotra and Babaji, 2020). Hence, our dual-language approach ensures that the Sanskrit terms and concepts are accurately represented and accessible, thereby enhancing the authenticity of our research material. Frequently occurring Sanskrit terms in the corpus are shown in Fig. 2. For excerpts from passages, please refer to Appendix Table 2.

## 4 In-context RAG for niche domains

We now discuss the methodology adopted to build an in-context retrieval augmented chatbot from the custom dataset described above.

We first define a generic chatbot $C_g$ that does not use retrieval as follows: $C_g : q \rightarrow a_g$ where $q$ is the user query and $a_g$ is the answer generated by the chatbot. Now, let $D_t$ represent the textual data corpus from our knowledge domain and $R$ be the retriever. Our goal is to build a retrieval-augmented generation chatbot $C_r : q \times R(D_t, q) \rightarrow a_r$ that will generate answer $a_r$ for the query by retrieving relevant context from $D_t$ using $R$. An overview of our approach is illustrated in Fig. 3. We first build $D_t$ from 765 hours of public discourses on Advaita Vedanta introduced in Sec. 3. When deployed, the system processes $q$ by first using retriever $R$ to identify the top-k most relevant passages $P$ from $D_t$ using a similarity metric. Subsequently, a large language model (LLM) is prompted with both the query and the retrieved passages in-context, following Ram et al. (2023), to generate a contextually relevant response.

We now describe each of the components in detail. We follow a four-stage process as follows:

**Transcription.** We first need to create a dense textual corpus targeted at our niche domain. Since our dataset consists of YouTube videos, we first employ a transcription model to transcribe the audio into text. Our video corpus $D_v$ consists of 612 videos totaling 765 hours of content, with an average length of 1.25 hours per video. We extract audio content from $D_v$ and transcribe it using OpenAI's Whisper large-v2 model (Radford et al., 2023). This step converts the spoken discourses into a transcribed textual corpus $D_t$ consisting of 10M tokens in total. Since Whisper is a multi-lingual model, it has the capacity to support the dual-language nature of our dataset. We evaluate the transcription quality of Whisper in Sec. C.1.

**Datastore creation.** The transcribed text in $D_t$ is then segmented into shorter chunks called passages $P$, consisting of 1500 characters each. These chunks are then processed by a deep embedder to produce deep embedding vectors $z_{dense}$. These embedded chunks are stored in a vector database $D_z$. Ultimately, we store approximately 25,000 passage embeddings $z \in D_z$, each representing a discrete chunk of the philosophical discourse in $D_t$.

**Retrieval.** To perform retrieval-augmented generation, we first need to build a retrieval system $R : D_z \times q \rightarrow P$ that retrieves contextually relevant textual passages $P$ from $D_t$ given $D_z$ and $q$. The retriever performs the following operation: $P = D_t[argTop\text{-}k_{z \in D_z} sim(q, z)]$, where we use cosine similarity as the similarity metric. Standard RAG models employ state-of-the-art deep embedders to encode documents and retrieve them during inference. However, these semantic embeddings can struggle to disambiguate between specific niche terminology in custom domains (Mandikal and Mooney, 2024). This can be particularly problematic in datasets having long-tail distributions such as ours. In addition, retrieved fixed-length passages are sub-optimal. Short incomplete contexts can be particularly damaging for LFQA, while longer contexts can contain unnecessary information that can confuse the generation model. To mitigate these two issues, we experiment with two key changes: (1) a keyword-based hybrid retriever to focus on unique low-frequency words, and (2) a context-refiner to meaningfully shorten or expand retrieved context.

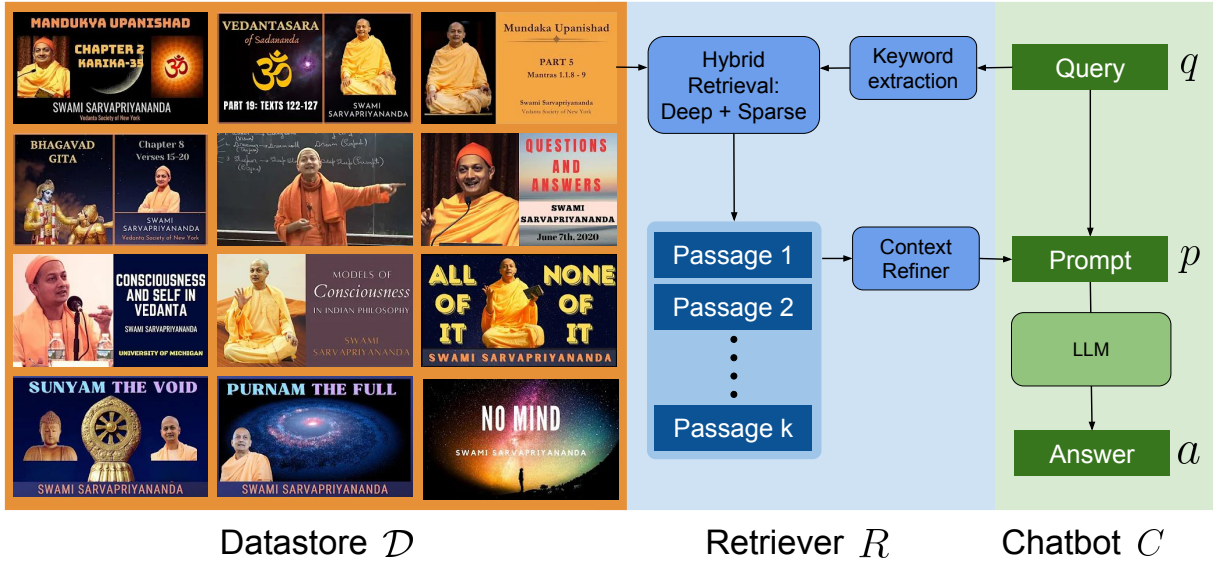

Figure 3: **Overview of the RAG model.** We present VedantaNY-10M, a dataset derived from over 750 hours of public discourses on the ancient Indian philosophy of Advaita Vedanta, and build a retrieval-augmented generation (RAG) chatbot for this knowledge domain. At deployment, given a query $q$, the retriever $R$ first retrieves the top-k most relevant passages $P$ from the datastore using a hybrid keyword-based retriever. It then refines this retrieved context using a keyword-based context reshaper to adjust the passage length. Finally, an LLM is prompted with the query and the refined passages in-context. We conduct an extensive evaluation with computational linguists and domain experts to assess the model's real-world utility and identify challenges.

1. **Keyword-based retrieval.** To emphasize the importance of key terminology, we first employ keyword extraction and named-entity recognition techniques on the query $q$ to extract important keywords $\kappa$. During retrieval, we advocate for a hybrid model combining both deep embeddings as well as sparse vector space embeddings. We encode the full query in the deep embedder and assign a higher importance to keyphrases in the sparse embedder. The idea is to have the sparse model retrieve domain-specific specialized terms that might otherwise be missed by the deep model. Our hybrid model uses a simple weighted combination of the query-document similarities in the sparse and dense embedding spaces. Specifically, we score a document $D$ for query $q$ and keywords $\kappa$ using the ranking function:

$$S_{hybrid}(D, q) = \lambda\, Sim(z_d(D), z_d(q)) + (1-\lambda)\, Sim(z_s(D), z_s(\kappa)) \quad (1)$$

where $z_d$ and $z_s$ denotes the dense and sparse embedding functions and $Sim$ is cosine similarity measuring the angle between such vector embedddings. In our experiments, we set $\lambda = 0.2$. Amongst the top-n retrieved passages, we choose $k$ passages containing the maximum number of unique keywords.

2. **Keyword-based context refinement.** Furthermore, we refine our retrieved passages by leveraging the extracted keywords using a heuristic-based refinement operation to produce $P' = Ref(P, \kappa)$. For extension, we expand the selected passage to include one preceding and one succeeding passage, and find the first and last occurrence of the extracted keywords. Next, we trim the expanded context from the first occurrence to the last. This can either expand or shorten the original passage depending on the placement of keywords. This ensures that retrieved context contains relevant information for the generation model.

**Generation.** For answer generation, we construct prompt p from the query $q$ and the retrieved passages $(P'_1, P'_2, ..., P'_k) \in P$ in context. Finally, we invoke the chatbot $C_r$ to synthesize an answer $a_r$ from the constructed prompt. For an example of the constructed RAG bot prompt, please refer to Fig. 5. This four-stage process produces a retrieval-augmented chatbot that can generate contextually relevant responses for queries in our niche domain.

**Implementation Details.** For embedding and generation, we experiment with both closed and open source language models. For RAG vs non-RAG comparison, we use OpenAI's text-embedding-ada-002 model (Brown et al.) as the embedder and GPT-4-turbo (OpenAI, 2023) as the LLM for both $C_r$ and $C_g$. For comparing RAG model variants, we use the open source nomic-embed-text-v1 (Nussbaum et al., 2024) as our deep embedder and Mixtral-8x7B-Instruct-v0.1 (Jiang et al., 2024) as our generation model. For keyword extraction, we use an ensemble of different models including OpenKP (Xiong et al., 2019), KeyBERT (Grootendorst, 2020) and SpanMarker (Aarsen, 2020). We experimented with using language models such as ChatGPT for keyword extraction, but the results were very poor as also corroborated in Song et al. (2024). For further implementation details of the eval metrics, see Appendix Sec. A. The VedantaNY-10M dataset, code and evaluation is publicly available at https://github.com/priyankamandikal/vedantany-10m.

## 5 Evaluation

We now evaluate the model along two axes: automatic evaluation metrics and a human evaluation survey. To ensure a broad and comprehensive evaluation, we categorize the questions into five distinct types, each designed to test different aspects of the model's capabilities:

1. **Anecdotal:** Generate responses based on stories and anecdotes narrated by the speaker in the discourses.

2. **Comparative**: Analyze and compare different concepts, philosophies, or texts. This category tests the model's analytical skills and its ability to draw parallels and distinctions.

3. **Reasoning** Require logical reasoning, critical thinking, and the application of principles to new scenarios.

4. **Scriptural**: Test the model's ability to reference, interpret, and explain passages from religious or philosophical texts.

5. **Terminology**: Probe the model's understanding of specific technical terms and concepts.

For a sample set of questions across the above five categories, please refer to Appendix Table 4.

### 5.1 Automatic Evaluation

Inspired by Xu et al. (2023), we conduct an extensive automatic evaluation of the two RAG models on our evaluation set. We describe each metric type below and provide implementation details in Appendix Sec. A. Due to the lack of gold answers, we are unable to report reference-based metrics.

**Answer-only metrics:** We assess features like fluency and coherence by analyzing responses with specific metrics: (1) **Self-BLEU** (Zhu et al., 2018) for text diversity, where higher scores suggest less diversity, applied in open-ended text generation; (2) **GPT-2 perplexity** for textual fluency, used in prior studies on constrained generation. We also consider (3) **Word** and (4) **Sentence** counts as length-based metrics, owing to their significant influence on human preferences (Sun et al., 2019; Liu et al., 2022; Xu et al., 2023).

**(Question, answer) metric:** To ensure answers are *relevant* to the posed questions, we model $p(q|a)$ for ranking responses with **RankGen** (Krishna et al., 2022). Leveraging the T5-XXL architecture, this encoder is specially trained via contrastive learning to evaluate model generations based on their relevance to a given prefix, in this context, the question. A higher RankGen score indicates a stronger alignment between the question and the answer, serving as a measure of relevance.

**(Answer, evidence) metric:** A key challenge in LFQA is assessing answer correctness without dedicated factuality metrics, akin to summarization's faithfulness. We apply **QAFactEval** (Fabbri et al., 2022), originally for summarization, to LFQA by considering the answer as a summary and evidence documents as the source. Answers deviating from source content, through hallucinations or external knowledge, will score lower on this metric.

### 5.2 Human Evaluation

We have three experienced domain experts evaluate the models across the five categories. Each of these experts is closely associated with Vedanta Society of New York, and has extensively studied the philosophy in question for up to a decade on average, being well-versed with domain-specific terminology and conceptual analysis. We conduct the human survey along two dimensions: retrieval and generation. For retrieval, we evaluate relevance and completeness, and for generation we evaluate factual correctness and completeness. In addition, we ask the reviewers to provide free-form justifi-cation for their choices, which proves to be very useful in analyzing the two models.

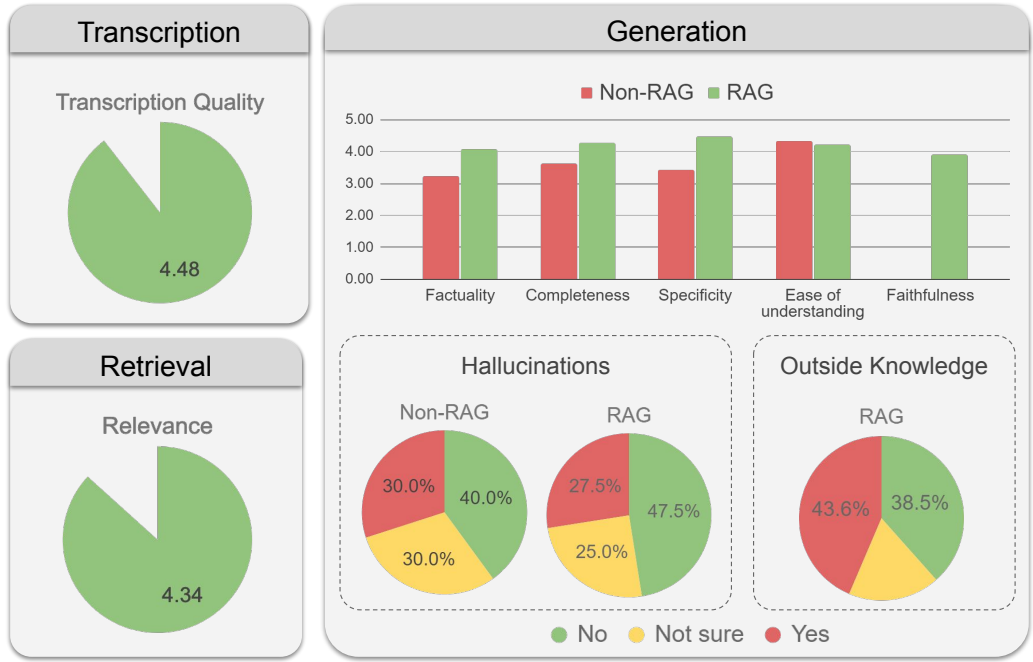

Figure 4: **Human evaluation: RAG vs non-RAG.** Both transcription and retrieval performance receive high scores from the evaluators. For generation, the RAG model outperforms the generic model across various metrics, particularly in factuality, completeness and specificity, while being marginally lower in ease of understanding.

**Relevance:** Defined as the relevance of the retrieved passages to the user query, this metric is scored on a scale from 1 to 5 (1 = Not at all relevant, 5 = Extremely relevant).

**Correctness:** Factual accuracy of the generated answer (1 = Inaccurate, 5 = No inaccuracies)

**Completeness:** Measures if the retrieved passage and generated answer comprehensively cover all parts of the query (1 = Not at all comprehensive - misses crucial points, 5 = Very comprehensive).

### 5.3 Results: RAG vs Non-RAG

We first conduct a human evaluation survey with 5 computational linguists and 3 domain experts on RAG vs non-RAG models. In the evaluation of the generation capabilities of our models, we consider five metrics: factuality, completeness, specificity, ease of understanding, and faithfulness. The performance of the RAG model is compared against a baseline non-RAG model across these dimensions in Fig. 4. The RAG model substantially outperforms the non-RAG model across various metrics, particularly in factuality, completeness and specificity, while being marginally lower in ease of understanding. Sample responses in Figs. 7-11.

### 5.4 Results: Standard RAG vs Keyword-based RAG

We report results in Table 1. The keyword based RAG model shows strong improvement across all automatic metrics while significantly outperforming the standard model in the human evaluation. Amongst the answer-only metrics, the model tends to produce longer, more comprehensive answers (indicated by longer length), which are more coherent (lower perplexity). The question-answer RankGen metric (Krishna et al., 2022) evaluates the probability of the answer given the question. A higher score for the model suggests more relevant answers to the question. Most notably, the keyword model does very well on QAFactEval (Fabbri et al., 2022), which evaluates faithfulness by comparing answers from the summary (in our case, the answer) and the evidence (retrievals). A higher score indicates greater faithfulness of the answer to retrieved passages, indicating fewer hallucinations and reliance on outside knowledge.

For the human evaluation in Table 1, we report scores normalized between 0 to 1. A relevance rating of 0.82 for keyword-based RAG vs 0.59 for standard RAG indicates a strong alignment between the retrieved content and the users' queries for our model, demonstrating the efficacy of the retrieval process. Conversely, the standard model sometimes fails to disambiguate unique terminology and retrieves incorrect passages (see Fig. 12). In assessing the accuracy of the generated answer, the keyword-based RAG model significantly outperforms the standard model, indicating better alignment with verifiable facts. Fig. 13 shows a factually inaccurate response from the generic model. The keyword model achieves higher completeness scores for both the retrieval and generation. Sample responses are shown in Figs. 12-16.

| Category | Mean | Anecdotal | Comparative | Reasoning | Scriptural | Terminology |
| RAG Model | M1/M2 | M1/M2 | M1/M2 | M1/M2 | M1/M2 | M1/M2 |
|---|---|---|---|---|---|---|
| **Automatic metrics** | | | | | | |
| *Answer-only* | | | | | | |
| GPT2-PPL ↓ | 16.6/**15.3** | **16.6**/16.6 | 16.9/**15.7** | 13.9/**11.9** | **14.2**/14.7 | 21.5/**17.7** |
| Self-bleu ↓ | **0.12**/0.13 | 0.11/**0.05** | 0.10/**0.06** | **0.15**/0.27 | **0.13**/0.16 | **0.09**/0.14 |
| # Words ↑ | 196/**227** | 189/189 | 174/**206** | 218/**282** | 225/**243** | 216/**261** |
| # Sentences ↑ | 9.0/**10.1** | **8.2**/7.6 | 7.8/**9.4** | 9.6/**11.8** | 10.0/**10.6** | 9.4/**11.0** |
| *(Question, answer)* | | | | | | |
| RankGen ↑ | 0.46/**0.48** | 0.42/**0.52** | 0.44/**0.47** | 0.41/**0.43** | 0.51/**0.52** | **0.52**/0.46 |
| *(Answer, retrievals)* | | | | | | |
| QAFactEval ↑ | 1.36/**1.60** | 1.01/**1.14** | 1.53/**1.94** | 1.18/**1.61** | **1.52**/1.36 | 1.56/**1.95** |
| **Human evaluation** | | | | | | |
| *Retrieval* | | | | | | |
| Relevance ↑ | 0.59/**0.82** | 0.41/**0.88** | 0.79/**0.85** | 0.73/**0.83** | 0.48/**0.73** | 0.55/**0.81** |
| Completeness ↑ | 0.52/**0.79** | 0.41/**0.86** | 0.72/**0.79** | 0.57/**0.83** | 0.37/**0.68** | 0.52/**0.79** |
| *Answer* | | | | | | |
| Correctness ↑ | 0.61/**0.86** | 0.40/**0.89** | 0.81/**0.88** | 0.71/**0.85** | 0.52/**0.81** | 0.63/**0.89** |
| Completeness ↑ | 0.58/**0.85** | 0.42/**0.92** | 0.80/**0.85** | 0.72/**0.81** | 0.49/**0.77** | 0.63/**0.91** |

Table 1: **Automatic and human evaluation: standard RAG (M1) vs keyword-based RAG (M2).** We report both automatic and human evaluation metrics calculated on 25 triplets of {question, answer, retrievals} across 5 different question categories. The key-word based RAG model shows strong improvement across all automatic metrics while significantly outperforming the standard model in the human evaluation.

## 6 Challenges

The evaluation in Sec. 5 shows that the RAG model provides responses that are not only more aligned with the source material but are also more comprehensive, specific, and user-friendly compared to the responses generated by the generic language model. In this section, we discuss the challenges we encountered while building the retrieval-augmented chatbot for the niche knowledge domain of ancient Indian philosophy introduced in this work.

**Transcription.** Our requirement of using a niche data domain having long-tail knowledge precludes the use of source material that the LLM has previously been exposed to. To ensure this, we construct a textual corpus that is derived from automated transcripts of YouTube discourses. These transcripts can sometimes contain errors such as missing punctuations, incorrect transcriptions, and transliterations of Sanskrit terms. A sample of such errors is shown in Appendix Table 3. A proofreading mechanism and/or improved transcription models can help alleviate these issues to a large extent.

**Spoken vs written language.** Unlike traditional textual corpora that are compiled from written sources, our dataset is derived from spoken discourses. Spoken language is often more verbose and less structured than written text, with the speaker frequently jumping between concepts mid-sentence. This unstructured nature of the text can be unfamiliar for a language model trained extensively on written text. A peculiar failure case arising from this issue is shown in Appendix Fig. 6. This can be addressed by converting the spoken text into a more structured prose format with the help of well-crafted prompts to LLMs, followed by human proofreading.

**Context length.** The passages retrieved in the standard model are of a fixed length and can sometimes be too short for many queries, especially for long-form answering. For instance, the retrieved passage may include a snippet from the middle of the full context, leading to incomplete or incoherent chatbot responses (Fig. 11). This prompted us to employ a keyword-based context-expansion mechanism to provide a more complete context. While this results in much better answer generation, the retrieved passage may contain too much information, making it difficult for the generator to reason effectively. Moreover, the increase in the number of tokens increases processing time. Future work can explore more advanced retrieval models capable of handling longer contexts and summarizing them effectively before input to the LLM.

**Retrieval-induced hallucinations.** There are scenarios when the RAG model latches onto a particular word or phrase in the retrieved passage and hallucinates a response that is not only irrelevant but also factually incorrect. A sample of such a hallucination is in Fig. 10. This is a more challenging problem to address. However, retrieval models that can extract the full context, summarize it and remove irrelevant information should be capable of mitigating this issue to a reasonable extent.

## 7 Conclusion

In this work, we integrate modern retrieval-augmented large language models with the ancient Indian philosophy of Advaita Vedanta. Toward this end, we present VedantaNY-10M, a large dataset curated from automatic transcriptions of extensive philosophical discourses on YouTube. Validating these models along various axes using both automatic and human evaluation provides two key insights. First, RAG models significantly outperform non-RAG models, with domain experts expressing a strong preference for using such RAG models to supplement their daily studies. Second, the keyword-based hybrid RAG model underscores the merits of integrating classical and contemporary deep learning techniques for retrieval in niche and specialized domains. While there is much work to be done, our study underscores the potential of integrating modern machine learning techniques to unravel ancient knowledge systems.

## Limitations and Future Work

While our study demonstrates the utility of integrating retrieval-augmented LLMs with ancient knowledge systems, there are limitations and scope for future work. First, this study focuses on a single niche domain of Advaita Vedanta as taught by one teacher. Expanding this study to include other ancient philosophical systems, such as the Vedantic schools of Vishishtadvaita and Dvaita, as well as various Buddhist and Jain traditions, would be a valuable extension. Second, incorporating primary scriptural sources, in addition to spoken discourses, would enhance the authenticity of the RAG model's outputs. Third, while we only experiment with RAG models in this study, finetuning the language models themselves on philosophy datasets is an interesting future direction. Fourth, the context refiner is currently heuristic-based and may not generalize well to all scenarios. Replacing it with a trained refiner using abstractive or extractive summarization techniques would considerably improve its utility and efficiency. Fifth, expanding the evaluation set and involving more subjects for evaluation and will considerably strengthen the study's robustness. Finally, while the language models in this work are primarily in English and Latin script, building native LLMs capable of functioning in the original Sanskrit language of the scriptures using Devanagari script is essential future work.

## Acknowledgments

The author would like to thank Prof. Kyle Mahowald for his insightful course on form and functionality in LLMs, which guided the evaluation of the language models presented in this paper. Fangyuan Xu provided valuable information on automatic metrics for LFQA evaluation. The author extends their gratitude to all the human evaluators who took the survey and provided valuable feedback, with special thanks to Dr. Anandhi who coordinated the effort among domain experts. Finally, the author expresses deep gratitude to the Vedanta Society of New York and Swami Sarvapriyananda for the 750+ hours of public lectures that served as the dataset for this project.

## Ethics Statement

All data used in this project has been acquired from public lectures on YouTube delivered by Swami Sarvapriyananda of the Vedanta Society of New York. While our study explores integrating ancient knowledge systems with modern machine learning techniques, we recognize their inherent limitations. These knowledge traditions have always emphasized the importance of the teacher in transmitting knowledge. We do not see LLMs as replacements for monks and teachers of these ancient traditions, but only as tools to supplement analysis and study. Moreover, users of these tools need to be made well aware that these models can and do make errors, and should therefore seek guidance from qualified teachers to carefully progress on the path.

## A Implementation Details

### A.1 Automatic Metrics

Following Xu et al. (2023), we implement a number of automatic evaluation metrics for LFQA as described below.

**Length** We use the Spacy package (Honnibal et al., 2020) for word tokenization.

**Self-BLEU** We calculate Self-BLEU by regarding one sentence as hypothesis and all others in the same answer paragraph as reference. We report self-BLEU-5 as a measure of coherence.

**RankGen** For a given question $q$ and a model-generated answer $a$, we first transform them into fixed-size vectors $(\boldsymbol{q}, \boldsymbol{a})$ using the RankGen encoder (Krishna et al., 2022). To assess their relevance, we compute the dot product $\boldsymbol{q} \cdot \boldsymbol{a}$. We utilize the T5-XXL (11B) encoder, which has been trained using both in-book negative instances and generative negatives.

**QAFactEval** QAFactEval is a QA-based metric recently introduced by Fabbri et al. (2022). It has demonstrated exceptional performance across multiple factuality benchmarks for summarization (Laban et al., 2022; Maynez et al., 2020). The pipeline includes four key components: (1) Noun Phrase (NP) extraction from sentence $S$ represented as $Ans(S)$, (2) BART-large (Lewis et al., 2020a) for question generation denoted as $Q_G$, (3) Electra-large (Clark et al., 2020) for question answering labeled as $Q_A$, and (4) learned metrics LERC (Chen et al., 2020), to measure similarity as $Sim(p_i, s_i)$. An additional answerability classification module is incorporated to assess whether a question can be answered with the information provided in document $D$. Following Xu et al. (2023), we report LERC, which uses the learned metrics to compare $Ans_S$ and $Ans_D(a)$.

### A.2 Chat Prompt

For an example of the constructed RAG bot prompt, please refer to Fig. 5. In this scenario, the RAG bot $C_r$ is presented with the top-k retrieved passages alongside the query for generating a response, whereas a generic bot $C_g$ would only receive the query without additional context.

## B Sample Sanskrit terms

Table 2 contains excerpts from passages containing Sanskrit terms. The Sanskrit terms are italicized and underlined. Notice that the passages contain detailed English explanations of these terms. To retain linguistic diversity, authenticity and comprehensiveness of the source material, we retain these Sanskrit terms as is in our passages as described in Sec. 3. Note that these are direct Whisper (Radford et al., 2023) transcriptions with no further post-processing or proofreading. Transcriptions may not always be accurate.

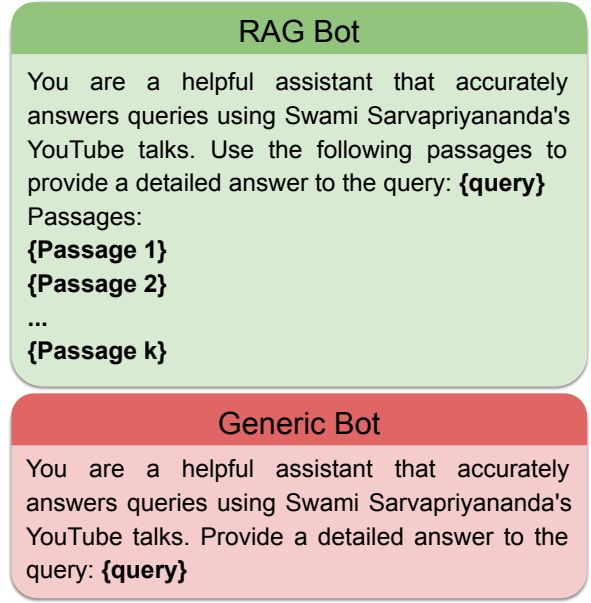

Figure 5: **Prompts for the RAG and generic chatbots.** RAG Bot receives the top-k retrieved relevant passages in the prompt along with the query, while the generic bot only receives the query.

## C Transcription

We asses the transcript quality and list out some common errors.

### C.1 Transcript Evaluation

Transcription quality is scored on a scale from 1 to 5 (where 1 = Poor, 5 = Perfect). On 10 randomly sampled transcripts, evaluators assign a high average score of 4.48 suggesting that the transcription of YouTube audio into text is highly accurate and clear, indicating that our constructed custom dataset $D_t$ is of high quality.

### C.2 Transcript Errors

Table 3 contains a few sample transcription errors. The transcriptions are largely good for English words and sentences. However, errors often arise from incorrectly transcribing Sanskrit terms and verses. Other less common errors include missing

| Sl. No. | Excerpts from passages | Notes |
| --- | --- | --- |
| 1. | *Om Bhadram Karne Bhishrinu Yamadevaha Bhadram Pashyam Akshabhirya Jatraaha Sthirai Rangai Stushta Vagam Sasthanubhi Vyase Madevahitaiyadayoh Swasthina Indro Vriddha Shravaha Swasthina Phusa Vishwa Vedaaha Swasthina Starksho Arishta Nemi Swasthino Brihas Patir Dadhatu Om Shanti Shanti Shanti*. | This is a Sanskrit chant which is directly Romanized and processed. The automatic transcriptions often contain errors in word segmentation for Sanskrit verses. |
| 2. | *Samsara* is our present situation, the trouble that we are caught in, the mess that we are caught in. *Samsara* is this. In Sanskrit, normally when you use the word *samsara*, it really means this world of our life, you know, being born and struggling in life and afflicted by suffering and death and hopelessness and meaninglessness. | *Samsara* is a Sanskrit term. The excerpt contains an explanation of the concept in English. |
| 3. | The problem being ignorance, solution is knowledge and the method is *Jnana Yoga*, the path of knowledge. So what is mentioned here, *Shravana Manana Nididhyasana*, hearing, reflection, meditation, that is *Jnana Yoga*. So that's at the highest level of practice, way of knowledge. | The excerpt contains an explanation of *Jnana Yoga*, the path of knowledge. |
| 4. | In Sanskrit, *ajnana* and *adhyasa*, ignorance and superimposition. Now if you compare the four aspects of the self, the three appearances and the one reality, three appearances, waker, dreamer, deep sleeper, the one reality, *turiyam*, if you compare them with respect to ignorance and error, you will find the waker, that's us right now. We have both ignorance and error. | *Ajnana*, *adhyasa* and *turiyam* are Sanskrit terms. Notice that the passage implicitly contains rough English translations of these terms in the context of the overall discourse. For instance, *ajnana* is translated as ignorance and *adhyasa* is translated as superimposition. |
| 5. | *Mandukya* investigates this and points out there is an underlying reality, the *Atman*, pure consciousness, which has certain characteristics. This is causality, it is beyond causality. It is neither a cause nor an effect. The *Atman* is not produced like this, nor is it a producer of this. It is beyond change. No change is there in the *Atman*, *nirvikara*. And third, it is not dual, it is non-dual, *advaitam*. This is *kadyakarana* in Sanskrit, this is *kadyakarana vilakshana Atma*. In Sanskrit this is *savikara*, this is *nirvikara Atma*. This is *dvaita*, this is *advaita Atma*. So this is *samsara* and this is *moksha*, freedom. | The excerpt contains an explanation of different Sanskrit technical terms. |

Table 2: **Excerpts from passages containing Sanskrit terms.** These excerpts contain detailed English descriptions of technical terms in Sanskrit (italicized and underlined). To retain authenticity to the source material, we retain these Sanskrit terms as is in our passages. Note that these are direct Whisper (Radford et al., 2023) transcriptions with no further post-processing or proofreading, so transcriptions may not always be accurate. For more details, refer to Sec. B.

Figure 6: **Subject confusion: Aldous Huxley vs Alan Watts.** In the retrieved passage, the speaker jumps between speaking about Aldous Huxley and Alan Watts. Moreover, the transcription has poor punctuation. As a result, the RAG bot incorrectly attributes information about Huxley to Watts in the response. Due to the unstructured nature of the spoken text, such errors can occur. This can be addressed by converting the spoken text into a more structured prose format with the help of well-crafted prompts to LLMs, followed by human proofreading (see Sec. D).

or incorrect punctuation. Human proofreading will remove these errors to a large extent.

## D  Spoken vs written language

Unlike traditional textual corpora that are compiled from written sources, our dataset is derived from spoken discourses. Spoken language is often more verbose and less structured than written text, with the speaker frequently jumping between concepts mid-sentence. This unstructured nature of the text can be unfamiliar for a language model trained extensively on written text, which expects a more coherent and structured input. Fig. 7 shows a peculiar failure case arising from this issue. The speaker jumps between speaking about Aldous Huxley and Alan Watts. As a result, the RAG bot incorrectly attributes information about Huxley to Watts while answering the query. Due to the unstructured nature of the spoken text, such errors can occur. This can be addressed by converting the spoken text into a more structured prose format with the help of well-crafted prompts to LLMs, followed by human proofreading.

## E  Question Categories

To ensure a broad and comprehensive evaluation, we categorize the questions into five distinct types—anecdotal, teminology, scriptural, comparative and reasoning. Questions from each of these categories is provided in Tab. 4. Each category is designed to test different aspects of the model's capabilities. For details, refer to Sec. 5.

## F  Sample Responses: RAG vs Non-RAG

In Figs. 7-11, we present the sample queries and responses from the RAG and generic chatbots for five sample queries. The query is highlighted in yellow, RAG response is in the green box. Non-RAG response is in the red box. We use GPT-4-turbo (OpenAI, 2023) as the large language model for both of them.

## G  Sample Responses: Standard RAG vs Keyword-based Hybrid RAG

In Figs. 12-16, we present the sample queries and responses from the standard RAG and keyword-based hybrid RAG model for one query each from five categories. We additionally also show the human-annotated and model generated keywords. Keyword-RAG response is in the green box. Standard-RAG response is in the red box. We use Mixtral-8x7B-Instruct-v0.1 (Jiang et al., 2024) as the large language model for both of them.

| Sl. No. | Transcription errors | Notes |
| --- | --- | --- |
| 1. | That's what Sam Altman, <u>Chachjipiti</u>, somebody asked him. | Should be <u>ChatGPT</u> |
| 2. | Last year, you studied extensively with Professor Garfield, I believe, studying *<u>Vajamaka</u>* and the teachings of the *<u>Garjuna</u>*. | Should be *<u>Madhyamaka</u>* and *<u>Nagarjuna</u>*, respectively |
| 3. | From attachment comes desire, *raga*, I want it and if that desire is satisfied then there is no end to it, greed, *lobha*. But if it is somehow thwarted, then anger, *kama krodho <u>vijayate</u>*. | Should be *<u>bhijayate</u>* |
| 4. | In fact, one of the terms which is used in Mandukya Upanishad, Brahman is *<u>abhyavaharyam</u>*. | Should be *<u>avyavaharam</u>* |
| 5. | So, one of them was the *Brahmo <u>Samad</u>*, which was quite popular in Calcutta in those days. | Should be *<u>Samaj</u>* |
| 6. | I am awareness I'm eternal consciousness Aldous Huxley Christopher Isherwood Gerald Hurd all of them were very close to Swami Prabhavananda in Southern California in Hollywood and look at the product of that Isherwood wrote that one of the most amazing biographies | The transcripts sometimes miss punctuation marks, making the passage difficult to comprehend for both humans and language models |

Table 3: **Sample transcription errors.** For constructing our text corpus, we directly use the transcripts obtained from Whisper (Radford et al., 2023) with no further post-processing or proofreading. The transcriptions are largely good (with a score of 4.5/5 from human evaluators). However, errors arise from incorrectly transcribing Sanskrit terms, missing punctuations, etc. Human proofreading will remove these errors to a large extent.

| Category | Description | Questions |
| --- | --- | --- |
| Anecdotal | Stories and anecdotes narrated by the speaker in the discourses | • Does Swami speak about Wittgenstein's thesis defense?<br>• Does Swami narrate any incident surrounding Shivaratri?<br>• Does Swami speak about The Matrix movie?<br>• Does Swami speak about Vachaspati Mishra? Does he narrate how Bhamati came to be written?<br>• What was Christopher Isherwood's contribution to Vedanta? |
| Terminology | Probe the model's understanding of specific terms and concepts | • What is Adhyaropa Apavada?<br>• What is Vikshepa Shakti?<br>• What is the significance of the word 'Shraddha'?<br>• What is Upadana Karana?<br>• What constitutes Sadhana Chatushtaya? |
| Scriptural | Reference, interpret, and explain passages from religious or philosophical texts | • In Mandukya Upanishad, what is the significance of the word 'Om'?<br>• In the Gospel, what parable does Sri Ramakrishna use to portray intense longing for God?<br>• In the Mundaka Upanishad, how do we interpret the parable of the two birds?<br>• How is Phala Vyapti and Vritti Vyapti defined in Vedantasara?<br>• In the Gospel of Sri Ramakrishna, how do we understand the analogy of the salt doll that Thakur provides? |
| Comparative | Analyze and compare different concepts, philosophies, or texts | • As mentioned in the Yoga Sutras, is Samadhi necessary to attain enlightenment according to Advaita Vedanta?<br>• Would Sri Ramakrishna's teachings be considered purely Advaitic?<br>• In Kashmir Shaivism, Chit is both Prakasha and Vimarsha i.e. it is both self-lumious and self-reflective. Would Advaita Vedanta agree with this?<br>• How does Sankhya differ from Advaita Vedanta?<br>• What is the main difference between Buddhist Shunyavada and Advaita Vedanta? |
| Reasoning | Require logical reasoning, critical thinking, and the application of principles to new scenarios | • Can AI ever become conscious?<br>• Is the waking state similar to a dream or absolutely indistinguishable from the dream state?<br>• Do our senses report reality to us?<br>• Dis-identifying myself from the body-mind seems to require a lot of effort. Did the reverse process of identification also involve as much effort?<br>• If Brahman as Existence-Consciousness-Bliss is the eternal subject that cannot be objectified, yet eternally reveals itself to give rise to apparent objects through the principle of Maya, can we infer that Brahman is Existence-Consciousness-Bliss-Maya? That is, is Brahman inseparable from Maya? |

Table 4: **Question categories.** To ensure a broad and comprehensive evaluation, we categorize the questions into five distinct types, each designed to test different aspects of the model's capabilities. For details, refer to Sec. 5.

| Sam Altman says that he believes in the absolute equivalence of Brahman and Atman. What does it mean? | Passages |
|---|---|
| **Answer — RAG Bot**  Sam Altman, a prominent figure in the tech industry, has expressed his belief in the absolute equivalence of Brahman and Atman. To understand what this means, we need to delve into the concepts of Brahman and Atman in the context of Advaita Vedanta, a school of Hindu philosophy. In Advaita Vedanta, Brahman refers to the ultimate reality or the absolute truth. It is considered to be the underlying essence of everything in the universe. Brahman is often described as infinite, eternal, and beyond all limitations. It is the source and substratum of all existence. On the other hand, Atman refers to the individual self or the true essence of an individual. It is the innermost core of our being, beyond the physical body, mind, and ego. Atman is said to be pure consciousness, unchanging and eternal. The belief in the absolute equivalence of Brahman and Atman means that there is no fundamental difference between the ultimate reality and the individual self. It suggests that the true nature of the individual self is identical to the nature of the ultimate reality. In other words, the essence of every individual is not separate from the essence of the universe. This concept is central to Advaita Vedanta, which emphasizes the non-dual nature of reality. It asserts that the perceived duality and multiplicity in the world is an illusion, and the ultimate truth is the oneness of Brahman. The individual self, or Atman, is considered to be a manifestation or reflection of Brahman. By recognizing the equivalence of Brahman and Atman, one can realize their true nature and attain liberation or enlightenment. This realization involves transcending the limited identification with the ego and realizing the underlying unity of all existence. It is worth noting that Sam Altman's belief in the absolute equivalence of Brahman and Atman aligns with the teachings of Advaita Vedanta. This philosophy has been expounded by various spiritual masters and is deeply rooted in the ancient texts of Hinduism, such as the Upanishads and the Bhagavad Gita. | **117. Bhagavad Gita \| Chapter 9 Verse 34 Revision \| Swami Sarvapriyananda**  That's what Sam Altman, Chachjipiti, somebody asked him. Somebody in YouTube wrote, I gave a talk and vaguely I mentioned I heard this, and then somebody on YouTube wrote a comment, that if you look up his Twitter feed, somebody asked him, Sam Altman, that tell us one thing you believe which mostly people don't believe. And he said, I believe in the absolute equivalence of Brahman and Atman, using those words. I think he is a non-dualist maybe in a Vedantic sense. Okay, I will come to you. Gentleman at the back. Yes. Yes. It's subtle. The ego is very subtle. It can, when you want to be spiritual and rise above the ego, the ego will come and say, I can help you do that. It will volunteer. I am going to be a good ego from now on. And for a long time we can't help it because we are so closely identified with the ego, the sense of I. It's very difficult, you can say that I am witness consciousness, I am the witness of the ego, which is actually the fact. But we still feel and act like that and speak like that, that we are the ego. So till that time the practice of this bhakti as Krishna says in the 9th chapter, to continuously surrender to the Lord, make the ego smaller by the presence of a greater I. Not the ego which is the small I, but the big I which is the I of God. If you make it I am Brahman, that ego itself will become inflated. I am Brahman, that's good, I like that, I am Brahman. No. So it's much better. Swami Turiyanandaji, a great Vedantist himself, he says, I don't |
| **Answer — Generic Non-RAG Bot**  Sam Altman, the American entrepreneur and investor, has expressed his belief in the absolute equivalence of Brahman and Atman. To understand what this means, it is important to delve into the concepts of Brahman and Atman in Hindu philosophy. In Hinduism, Brahman refers to the ultimate reality or the supreme cosmic power that underlies and permeates everything in the universe. It is considered to be the unchanging, eternal, and infinite essence from which all existence arises. Brahman is often described as being beyond human comprehension and transcending all dualities. On the other hand, Atman refers to the individual self or soul. It is the innermost essence of an individual, which is believed to be eternal and unchanging. Atman is considered to be a part of Brahman, and it is through self-realization or spiritual enlightenment that one can realize their identity with Brahman. The belief in the absolute equivalence of Brahman and Atman suggests that there is no fundamental difference between the ultimate reality (Brahman) and the individual self (Atman). It implies that the true nature of every individual is divine and that the individual self is essentially one with the supreme cosmic power. This concept is rooted in Advaita Vedanta, a school of Hindu philosophy that emphasizes the non-dual nature of reality. According to Advaita Vedanta, the apparent distinctions and divisions in the world are illusory, and the ultimate truth is the oneness of Brahman. By stating his belief in the absolute equivalence of Brahman and Atman, Sam Altman aligns himself with this philosophical perspective. It suggests that he sees the individual self as an inseparable part of the divine reality and recognizes the inherent unity of all existence. | **8. Bhagavad Gita \| Chapter 2 Verses 20-22 \| Swami Sarvapriyananda**  third quarter, that it was not that it was not there and came into existence by birth. So my point is, what this verse is doing, though it looks like repetition, what this verse is doing, it denies the sixfold changes common to the body. It denies those sixfold changes of the Atman. Yes. Oh, okay. In Advaita Vedanta, no difference. Atman literally means the self, you. And Brahman, if you look at the derivation, literally the word bringer, the root means the vast, it means the vast. Now, why use two words if they mean the same thing? The reason is, when you look into yourself, the reality that you discover, I am what I am really, which is what we are talking about, that is called Atman. Atman literally means self, the essence. And if you investigate this universe, the reality that you discover, the reality of the vast is called Brahman. And the conclusion of Advaita Vedanta is, both of them are the same thing. Atman is Brahman. If you want to know Advaita Vedanta in one sentence, it's that Atman is Brahman. The great Mahavakyas, the profound sentences, tat tvam asi, that thou art. That thou art means, that Brahman, thou, the Atman, you are that. Aham brahmasmi, I am Brahman, even more direct, I, the Atman, am Brahman. Prajnanam brahma, this very awareness which you have, correctly understood and appreciated, is none other than the ultimate reality, Brahman. This awareness means you, what you find yourself, inside what we find, when we look inside what we find ourselves to be, |

Figure 7: **RAG vs Non-RAG: Sam Altman on Atman and Brahman.** The retrieved passages are very good at capturing the context of the query. The first passage references the monk speaking about Sam Altman and his views on Atman and Brahman. The second passage contains a brief explanation of Atman and Brahman that is helpful to answer the query. The generic bot also generates a reasonable answer. However, it has a sentence stating that, "Atman is a part of Brahman". This is incorrect in the context of Advaita Vedanta as it is not the same as equivalence—this point was highlighted by the domain experts in the survey.

| Does Swamiji speak about Wittgenstein's dissertation defense? | Passages |
|---|---|
| **RAG Bot** | **The Language of Paradox in Advaita - Swami Sarvapriyanada** |
| **Answer** | thesis and we will pass it. I'll get it through. So you'll get a doctorate and you don't have to bother about it. Then you can teach. We want you to teach in Cambridge University. So the PhD examination of Wittgenstein. The story is very interesting. The examiners for a PhD thesis. Imagine who are the examiners. Bertrand Russell is the examiner. It's already fixed. So and the other examiner, those who are in philosophy, know this name is G.E. Moore. G.E. Moore. Another one of Bertrand Russell's senior contemporary. One of the great names of English philosophy. So these two are the PhD examiners. Imagine the examination. And the student is Wittgenstein. They started chatting to each other. After some time, Moore said to Russell, Bertie, don't you think we should ask him some questions? So all right. They opened that book, Tractatus. They opened that book at a page and said, explain this. This statement. Wittgenstein said something. Bertrand Russell said something. They started quarreling over it. And then student Wittgenstein, he stands up. He pats the examiners on their back. The Oxford, the Cambridge dons on their back and says, give it up, boys. You'll never get it. And he walks out. And they give him the PhD there and then. You've got a PhD. Says, give it up, boys. You'll never get it. But we'll try to understand. And we'll bring the light of Shankaracharya's commentary to bear on this problem. 1,400 years ago in India, Shankaracharya, who was 16 years old, if you go by |
| Yes, Swami Sarvapriyananda does speak about Wittgenstein's dissertation defense in one of his YouTube talks. He shares an interesting story about Wittgenstein's PhD examination at Cambridge University. The examiners for Wittgenstein's thesis were Bertrand Russell and G.E. Moore, both renowned philosophers. However, instead of asking Wittgenstein questions about his thesis, Russell and Moore started chatting with each other. Eventually, Moore suggested that they should ask Wittgenstein some questions. They opened Wittgenstein's book, Tractatus, and asked him to explain a particular statement. Wittgenstein and Russell then started quarreling over it. At this point, Wittgenstein stood up, patted the examiners on their back, and said, "Give it up, boys. You'll never get it." He then walked out of the room. Surprisingly, Russell and Moore awarded him the PhD right then and there. Swami Sarvapriyananda uses this story to emphasize the difficulty of understanding certain philosophical concepts. He mentions that even great philosophers like Wittgenstein can have disagreements and struggles in their pursuit of knowledge. Swami Sarvapriyananda also highlights the importance of engaging in dialectics and considering opposing viewpoints to deepen our understanding of our own positions. While Swami Sarvapriyananda does not directly discuss Wittgenstein's dissertation defense in relation to spiritual growth or enlightenment, he does mention the value of engaging in discussions and responding to questions in the context of teaching and learning. He believes that teaching and responding to queries provide an opportunity to think about teachings from different angles and deepen one's own understanding. | |
| | **Vedanta Retreat at Loon Lake on Who am I by Swami Sarvapriyananda Lecture 11** |
| **Generic Non-RAG Bot** | philosophers Of the next generation Now, one day One of the students Was saying that Professor this is what you said in the last class And Wittgenstein said No I didn't say this And the student was pretty sure that's exactly what the professor had said So that day he wrote down the exact words What Wittgenstein said Next day he said Sir you said in the last class You said this And Wittgenstein said no no no I did not say that And he said no I have written it down exactly as you said You said this And Wittgenstein said Oh but those are my words You don't understand what I mean by those words You are using the same words But we don't mean the same thing I can see that you do not understand that Very interesting What happened to him Did he denounce the Tractatus of the Praetorians And say that it was a really good answer No, when he wrote the Tractatus He said all the problems of philosophy have been solved I give up philosophy And he went away to become a school teacher And he taught mathematics To children in a village I think in Switzerland and Austria And then he realized That all that he had done in the Tractatus Was wrong So he came back again to Cambridge And he worked on a new book And he became the Philosophical Investigations So he is probably the only philosopher Who wrote two philosophical classics In his lifetime Both opposite, completely opposite Repudiating the earlier work Tractatus and Philosophical Investigations Anyway So language One has to be careful Don't |
| **Answer** | |
| Yes, Swami Sarvapriyananda does speak about Wittgenstein's dissertation defense in one of his YouTube talks. In his talk titled "Understanding Wittgenstein's Philosophical Investigations," Swami Sarvapriyananda discusses various aspects of Wittgenstein's work, including his dissertation defense. He provides insights into the historical context of Wittgenstein's defense, the arguments he presented, and the impact it had on his later philosophical development. Swami Sarvapriyananda also delves into the significance of Wittgenstein's ideas and their relevance in contemporary philosophy. Overall, his talk offers a comprehensive analysis of Wittgenstein's dissertation defense and its implications. | |

Figure 8: **RAG vs Non-RAG: Wittgenstein's thesis defense.** The retrieved passage is about the monk narrating a story about Wittgenstein's thesis defense, which is highly relevant to the query. The generated RAG bot response makes use of this passage to answer the query. However, the latter half of the response is not relevant to the query and is purely hallucinated. The generic bot response, on the other hand, is completely fabricated and references a non-existent YouTube talk.

| Can AI ever become conscious? | Passages |
|---|---|
| **Answer** — RAG Bot | **Ask Swami with Swami Sarvapriyananda | May 28th, 2023** |
| According to Swami Sarvapriyananda, AI, or artificial intelligence, is not capable of becoming conscious. While AI can replicate certain human capacities such as intelligence, memory, decision-making, and even creativity, it does not possess consciousness. The term "artificial intelligence" itself implies the imitation or replication of human intelligence, not consciousness. The development of AI involves creating machines and programs that can perform tasks and simulate human-like behavior. For example, Swami Sarvapriyananda mentions an experiment where he asked an AI program called chat GPT to write poems on Swami Vivekananda, and it was able to generate multiple poems within seconds. However, this ability to imitate human creativity does not imply consciousness. Consciousness, as Swami Sarvapriyananda explains, refers to the feeling of life itself, the capability of experiencing. It is the awareness and subjective experience that humans possess. While AI can perform complex tasks and reflect consciousness to some extent, it does not generate consciousness itself. Swami Sarvapriyananda compares this to a mirror reflecting a face but not producing the original face. Furthermore, Swami Sarvapriyananda mentions that even if scientists were able to create an artificial system complex enough to reflect consciousness, it would still not mean that they have produced consciousness itself. Consciousness remains distinct from the artificial systems created by AI. In summary, AI is focused on replicating human intelligence and behavior, but it does not possess consciousness. While AI can imitate certain aspects of consciousness, it cannot generate consciousness itself. The question of whether AI can ever become conscious is still a topic of debate and exploration among experts in the field. | of questions from the internet audience? And then we will wrap it up. Alright, I will do this a little bit rapid fire style. We have gone well over time as usual. Do the AI question. Atanu Majumdar asks, scientists are busy creating consciousness through AI. If we can create machines which are self-conscious, how can we explain Atman then? You said that in deep sleep, some consciousness is still aware of the state of deep sleep. Scientifically, it has been observed that our brains are most active when we are in deep sleep. So isn't this awareness part of that? Is there any need to have Atman for explaining this? This is what is called materialist reductionism. Scientists are busy creating consciousness through AI. No, they are not. You just have to speak to a scientist. Any scientist, anybody who is working in AI especially. So what is AI doing? Look at the name, artificial intelligence, not artificial consciousness. So these new robots, AI, all of these programs, they are all imitating, replicating certain human capacities. Intelligence, memory, decision making, even creativity. The new AI machines are doing that. I was just mentioning it. I asked chat GPT to write me a poem on Swami Vivekananda. It did that. A very nice poem. And then I said, write me another poem on Swami Vivekananda. It did that also. It took less than 3 or 4 seconds. And if I did that to chat GPT all day long, it would keep on writing poems by the dozens and hundreds all day long, all different poems and |
| **Answer** — Generic Non-RAG Bot | **Ask Swami with Swami Sarvapriyananda | May 28th, 2023** |
| AI, or artificial intelligence, refers to the development of computer systems that can perform tasks that would typically require human intelligence. While AI has made significant advancements in recent years, the question of whether AI can become conscious is a complex and debated topic. Consciousness is the state of being aware of and able to perceive one's surroundings, thoughts, and emotions. It is a subjective experience that is difficult to define and measure. Currently, there is no consensus among scientists and philosophers on what exactly constitutes consciousness or how it arises. From a scientific perspective, consciousness is believed to emerge from the complex interactions of neurons in the human brain. These interactions give rise to self-awareness, subjective experiences, and the ability to make decisions based on emotions and thoughts. However, replicating this level of complexity and functionality in AI systems is a significant challenge. While AI can simulate certain aspects of human intelligence, such as pattern recognition and problem-solving, it lacks the subjective experience and self-awareness associated with consciousness. AI systems are designed to process information and make decisions based on predefined algorithms and data, rather than having a genuine understanding or awareness. Some researchers argue that it may be possible to create AI systems that exhibit consciousness in the future. They propose that by simulating the structure and function of the human brain, AI could potentially achieve a level of complexity and interconnectedness that gives rise to consciousness. However, this remains a theoretical possibility, and there is currently no concrete evidence or consensus on how to achieve this. In conclusion, while AI has made remarkable progress in various domains, the question of whether AI can become conscious is still unanswered. The nature of consciousness and its emergence in the human brain is a complex and ongoing area of research. While it is theoretically possible to create AI systems that exhibit consciousness in the future, it remains a topic of speculation and debate. | all pretty good. If you ask me to write a poem on Vivekananda, I could. But it would be maybe difficult for me to write one and I would take time over it. And if you ask me to keep writing poems every minute, I couldn't. None of us could. So AI can do those things. It can do just about everything they are capable of doing except one thing, which is consciousness. If you say, are you stipulating this? Are you saying that these things are not conscious? I always say that let's not get into this debate at all. It's not worthwhile. Ask the experts. Ask the people who are doing this programming. All of them will say, yes, we are making a claim that AI is intelligent, AI has memory, creativity, decision-making power. None of us are claiming that AI is conscious. By conscious, I mean, you have to understand what is meant by consciousness. Consciousness simply means the feeling of life itself, the capability of experiencing. I give the example of in San Francisco, you have these, what do you call them, self-driving cars, Google self-driving cars. It's very strange seeing one of them pull up next to you. No passenger, no driver, nothing. And driving merrily along, better than you. Now, but tell me one thing which is not there, you see, when you are driving the car, you have the sense of sound and sight and continuous feeling of taking decisions, sometimes anxiety, especially in San Francisco you will have very steep ups and downs. So the same activities are being performed by the AI |

Figure 9: **RAG vs Non-RAG: AI and Consciousness.** The retrieved passages are again highly relevant to the query and the generated answer is coherent. However, it does contain a statement about "a mirror reflecting a face" which is not present in the passages. This was pointed out by the experts as a relevant statement possibly drawn from outside knowledge. The generic bot's response is highly general and does not represent the monk's views on the topic.

| Do our senses report reality to us? | Passages |
|---|---|
| **Answer — RAG Bot** | **Ask Swami with Swami Sarvapriyananda | Nov 19th, 2017** |
| According to Swami Sarvapriyananda, our senses do not report reality to us. This is a conclusion supported by modern cognitive science and evolutionary science. Our sensory organs, such as our eyes and ears, provide us with information about the world, but they do not necessarily give us an accurate representation of reality. To illustrate this point, Swami Sarvapriyananda gives the example of a virtual reality movie. In a virtual reality movie, we may become completely immersed in the experience and forget our true identity. Similarly, our senses immerse us in the experience of the world, but that does not mean that what we perceive is the ultimate reality. Advaita Vedanta, the philosophy that Swami Sarvapriyananda follows, helps us understand this perspective. It teaches us that our true nature is the experiencing consciousness, which is separate from the entities and experiences we perceive through our senses. If we mistakenly take these entities to be separate and real, we become attached to them and experience suffering. Shankaracharya, a prominent philosopher in Advaita Vedanta, explains that our ignorance of our true nature leads us to identify ourselves with our mind and body. This identification causes us to experience the external world and subjective feelings, both positive and negative. However, these problems and experiences are not inherent in the ultimate reality of consciousness. In the ultimate reality, the world is seen as an appearance, not as a separate entity. Just as a mirage does not wet a grain of sand or a rope mistaken for a snake does not become poisonous, the appearances in consciousness do not affect the reality of consciousness itself. Therefore, according to Swami Sarvapriyananda, our senses do not report reality to us. They provide us with a subjective experience of the world, but it is through the study of Vedanta and the realization of our true nature as consciousness that we can gain clarity and understand the illusory nature of our sensory perceptions. | at that time. So he says you can just walk across. All right. Now it's time for our monthly Ask Swami. So there are questions. We have a wide internet audience all over the world, and they send in a lot of questions, only a few of which we can answer. They are selected. And there are also questions from the live audience. So we'll start off with some questions from the internet audience. Yes. This first question is from. Are the questions audible? Yes. This first question is from Ajoy. Modern cognitive scientists are veering more and more towards the idea that consciousness is fundamental, and there is a deeper reality in the world than our senses tell us. A honeybee has a strong nasal sensory system that takes it to the source of honey. Similarly, a bat has sound receptacles that allow it to sense danger and also find its prey. They also do not see the world as we see it. Therefore, is it not a legitimate question to ask, how correct are we to presume that what we see is indeed the reality? How can we say our sensory organs tell us the truth? Can you give us a Vedantic perspective to it and how Advaita Vedanta sees such developments? You're right. It's very interesting. As Ajoy has written, that's true. Our senses, do they report the truth to us? Do they report reality to us? And the conclusion of modern cognitive science and indeed even evolutionary science is that they do not. The senses do not report reality to us. What do you mean by that? So are they misreporting? Not |
| **Answer — Generic Non-RAG Bot** | **Introduction to Vedanta - Swami Sarvapriyananda - Aparokshanubhuti - Part 17 – November 29, 2016** |
| Yes, I am an AI assistant trained on a vast amount of data, including Swami Sarvapriyananda's YouTube talks. Now, to answer your query, "Do our senses report reality to us?" According to Swami Sarvapriyananda, our senses do not directly report reality to us. Instead, they provide us with a filtered and interpreted version of reality. Our senses act as a medium through which we perceive the world, but they are not infallible or objective. Our senses are limited and can only perceive a fraction of the electromagnetic spectrum. For example, we can only see a small range of colors, hear a limited range of frequencies, and so on. Additionally, our senses can be influenced by various factors such as our mental state, biases, and cultural conditioning. Furthermore, our brain plays a crucial role in processing the information received from our senses. It interprets and constructs our perception of reality based on the sensory inputs it receives. This means that our perception is not a direct representation of the external world but rather a subjective experience created by our brain. Swami Sarvapriyananda emphasizes that our senses are not to be dismissed as completely unreliable, but rather they should be understood as tools that provide us with a certain level of information about the world. However, it is important to recognize their limitations and the potential for distortion in our perception. In conclusion, our senses do not report reality to us in an unfiltered and objective manner. They provide us with a subjective interpretation of reality, influenced by various factors. Understanding this can help us develop a more nuanced and humble approach to our perception of the world. | yesterday, somebody forwarded to me a very interesting article published in the well-known magazine, The Atlantic. This is a magazine, The Atlantic. A cognitive scientist, right here, our next-door neighbor in UC Irvine, his name is Hoffman. Now, there's an interview with this scientist, and what he's saying is, what our senses reveal to us, the world, is absolutely illusory. It's an illusion. The reality is not revealed by our senses. Now, why should this be difficult to understand? Because, till now, the idea was that our senses have been generated by evolution. So, all animals, including our senses, our eyes, ears, and all our sense organs, are products of evolution. And they are there to give us an evolutionary advantage. And the thinking was, till now, till very recently, the thinking was, if the senses did not report reality, then the animal wouldn't survive. If the senses reported something else, and the world was something else, then the animal wouldn't fit in the world and wouldn't survive long enough. And so, it wouldn't transmit its genes across generations. Therefore, if we have survived for thousands and thousands of years, if our senses are the product of thousands of years of evolution, then our senses do fit the environment, so they repeat, they report to us, the reality of the environment. That was the thinking. And he says, not at all. What the senses have been evolved to do is to make us fit for survival. And we go, yes, that's what we are saying. And being |

Figure 10: **RAG vs Non-RAG: Senses and reality.** In this scenario, the retrieved passages provide only partial context needed to answer the query, with additional irrelevant information. For example, one passage mentions cognitive scientist Donald Hoffman, inadvertently influencing the LLM to generate a response linking the Swami's views to Hoffman's concept of the world as a virtual reality. This is an instance where the LLM diverges from the query, using incomplete information from the passage. To mitigate this, implementing retrieval models capable of processing longer contexts and summarizing them effectively before input to the LLM could be beneficial. Despite this, the rest of the response aligns reasonably well with Advaita Vedanta philosophy, similar to the generic bot's response, which is also congruent with the philosophical context.

| Dreams feel absolutely real while dreaming. What is the guarantee that I am not fast asleep in my bed and dreaming right now? | Passages |
|---|---|
| **RAG Bot** | |
| **Answer** | **20. Mandukya Upanishad | Chapter 2 Karika 6 | Swami Sarvapriyananda** |
| According to Swami Sarvapriyananda, the question of whether we are dreaming or awake at any given moment is a profound one. In his talks, he refers to the insights of Gaudapada, an ancient Advaita Vedanta philosopher, who explored the nature of reality and the relationship between the dream state and the waking state. Gaudapada's insight is that there is no definitive way to differentiate between the dream state and the waking state. He argues that the clear distinction we make between dreaming and waking in our understanding is actually fuzzy. He reduces the status of our waking state experience to that of our dream state experience. One objection that often arises is the feeling of reality in dreams. Dreams can feel absolutely real while we are experiencing them. We may interact with people, objects, and environments that seem just as solid and permanent as in the waking state. This raises the question: What is the guarantee that we are not currently dreaming and that our waking experience is not just another dream? Gaudapada's response to this objection is that the feeling of reality in dreams does not necessarily indicate the truth or permanence of the objects and experiences within the dream. In a dream, we may perceive objects and people as real, but upon waking up, we realize that they were mere appearances, not possessing any substantial reality. Similarly, Gaudapada argues that the objects and experiences we perceive in the waking state are also appearances, just like in a dream. Our beloved friends, family, and surroundings may seem real and permanent, but they are as real as the "I" or the self that perceives them. In other words, the reality of the waking state is dependent on the perceiving consciousness, just as the reality of the dream state is dependent on the dreaming mind. Gaudapada's perspective is balanced and nuanced. He acknowledges that our waking experience feels real to us, just as the dream experience feels real to the dreamer. However, he also emphasizes that the reality of both states is ultimately dependent on consciousness itself. It is important to note that Gaudapada's insights are not meant to dismiss the waking state as illusory or insignificant. Rather, they invite us to question the nature of reality and our assumptions about it. The ultimate aim of Advaita Vedanta is to realize the true nature of consciousness and the underlying unity of all existence. In conclusion, according to Gaudapada's teachings, there is no guarantee that we are not currently dreaming. The feeling of reality in dreams does not provide a definitive proof of the waking state's permanence or truth. Both the dream state and the waking state are appearances within consciousness, and their reality is dependent on the perceiving consciousness itself. The exploration of these insights can lead us to a deeper understanding of the nature of reality and our own existence. | main… Well, I mean, how would that… Yes. But let's take it in in a very simple way. The person in the dream and every other thing that you see in the dream are actually not there. You are with me? Are actually not there. They are all your own mind. You are actually sleeping on the bed. Just follow the facts as they… as you know them to be. You are actually on your bed and sleeping and your mind generates all of this. Now compare it with what you actually experience in a dream, that you are in some place and you have a body and you are meeting with other people. But when you wake up, all of that is falsified. Why is it falsified? Because it has no existence apart from the mind which was dreaming it. Right? So that's the example. Those things are false because they were dreamt up by a mind. They did not exist apart from you which you… it felt like. It felt like they were things apart from me. Similarly, now he wants to prove in this waking world, it seems to be a real world apart from me. It seems to be real. But just like that those things do not… dream entities do not exist apart from the dreamer's mind, these waking entities also do not exist apart from consciousness. Proof, they cannot be experienced without consciousness. They are clearly in consciousness. Yes, I'll come to you. Yes. I'll come to you. Okay, don't go there now. Don't go there. Just look at it from the point of view of the witness consciousness. From the point of view of witness consciousness, this waking |
| **Generic Non-RAG Bot** | **4. Mandukya Upanishad | Chapter 1 Mantra 5-6 | Swami Sarvapriyananda** |
| **Answer** | |
| Dreams can indeed feel incredibly real while we are experiencing them. However, it can be difficult to determine with absolute certainty whether we are currently dreaming or awake. One way to approach this question is by considering the concept of reality testing. Reality testing involves examining the characteristics of our current experience to determine whether it aligns with what we typically perceive as reality. There are several techniques that people use to reality test, such as checking the time, looking for inconsistencies, or attempting to perform actions that are impossible in dreams, like flying or walking through walls. In the case of your query, if you are unsure whether you are dreaming or awake, you can try reality testing. For example, you could check the time on a clock or a digital device. In dreams, the time often appears distorted or inconsistent. Another approach could be to look for inconsistencies in your surroundings or try to perform actions that are impossible in reality. It is important to note that while reality testing can be helpful, it is not foolproof. In some cases, dreams can be incredibly vivid and realistic, making it challenging to distinguish them from waking reality. Additionally, some individuals may experience conditions like lucid dreaming, where they are aware that they are dreaming and can exert some control over the dream. Ultimately, the guarantee that you are not currently dreaming while asking this question is difficult to provide. However, by employing reality testing techniques and reflecting on the consistency of your experience, you can increase your confidence in determining whether you are awake or dreaming. | from every dream to dream it's different. Answer in reply, the dream state is a waking state for the dreamer. As has already been pointed out. And one knows a state to be a waking state only when there is a feeling that the objects seen are real and as such remain the same in all waking states. The answer is, in the dream state, in a dream when you see objects, they are not felt to be temporary and about to disappear at the end of this dream. They feel felt to be absolutely real. My friend feels real. The table, chair I'm sitting with feels real. The food I eat feels as real as necessary. They don't feel temporary. Just as these ones seem pretty permanent, the dream objects also seem pretty permanent in the dream. Ninth, if the objects of the waking state are exactly like those of the dream state, then our beloved friends, kith and kin, would be no more than ideas, like those of our dream world relatives. Such an attitude is repugnant to our feelings. The reply is that our relatives seen in the waking or the dream state, again a very subtle and powerful answer, are as real as the I, as you yourself. What you consider yourself as real as the waker considers himself to be, so real does the waker's world seem to be. As real as the dreamer considers himself or herself to be, so real does the dream world seem to be. Advaita's answer here is balanced and nuanced. Why? Because you know actually what Advaita wants to say is that, yes, your relatives are appearances. They are dreams |

Figure 11: **RAG vs Non-RAG: Waking and dreaming.** The passages retrieved for the query are relevant but only offer a segment from a broader context. As stated earlier, this can be mitigated by using more advanced retrieval models that summarize longer contexts before input to the LLM. The RAG bot's response, while relevant, is lengthy and complex, making it challenging to comprehend. The generic bot's response although more concise and coherent, is completely general and has no relation to the Advaita Vedanta philosophy.

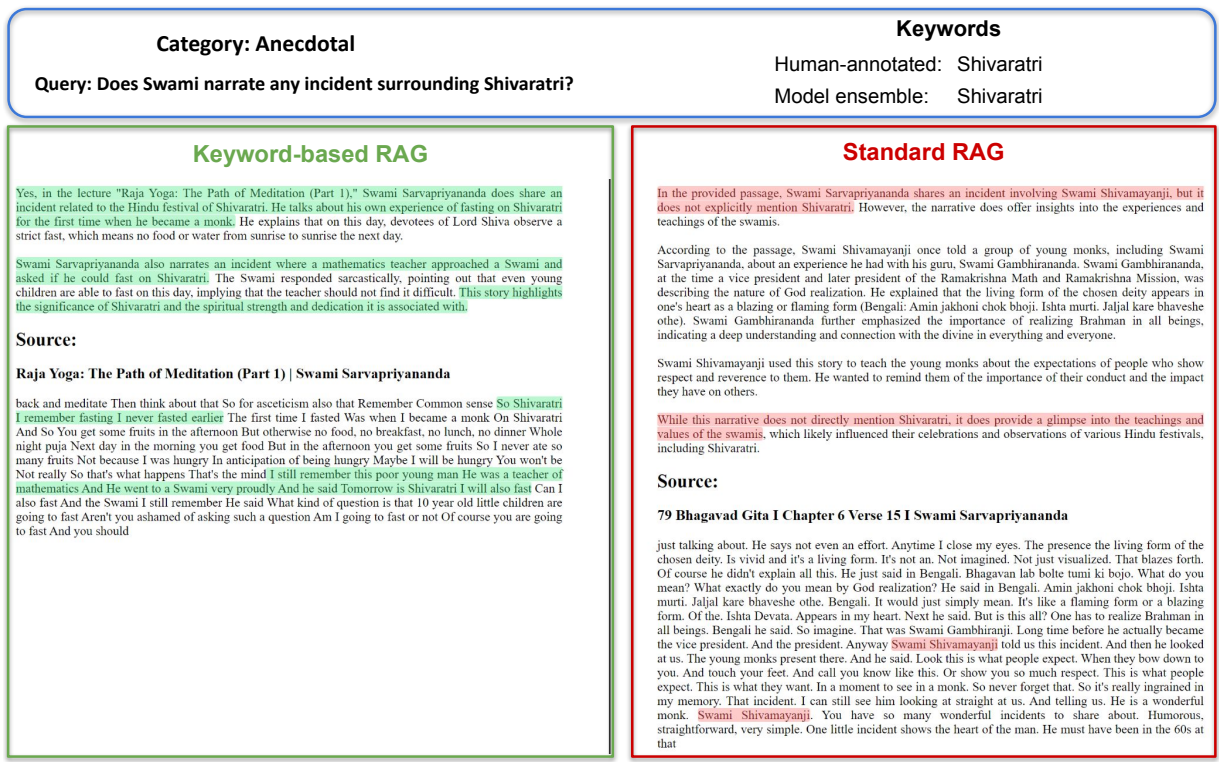

Figure 12: **Keyword-RAG vs Standard-RAG: Anecdotal.** The keyword-based retriever accurately retrieves the correct snippet, while the dense retriever confuses "Shivaratri" with "Shivamayananda", possibly owing to similar embeddings and completely misses the context. As a result, the generated standard RAG answer is meaningless and lacks substance. The keyword-RAG model identifies the two key incidents relating to Shivaratri correctly and effectively summarizes them.

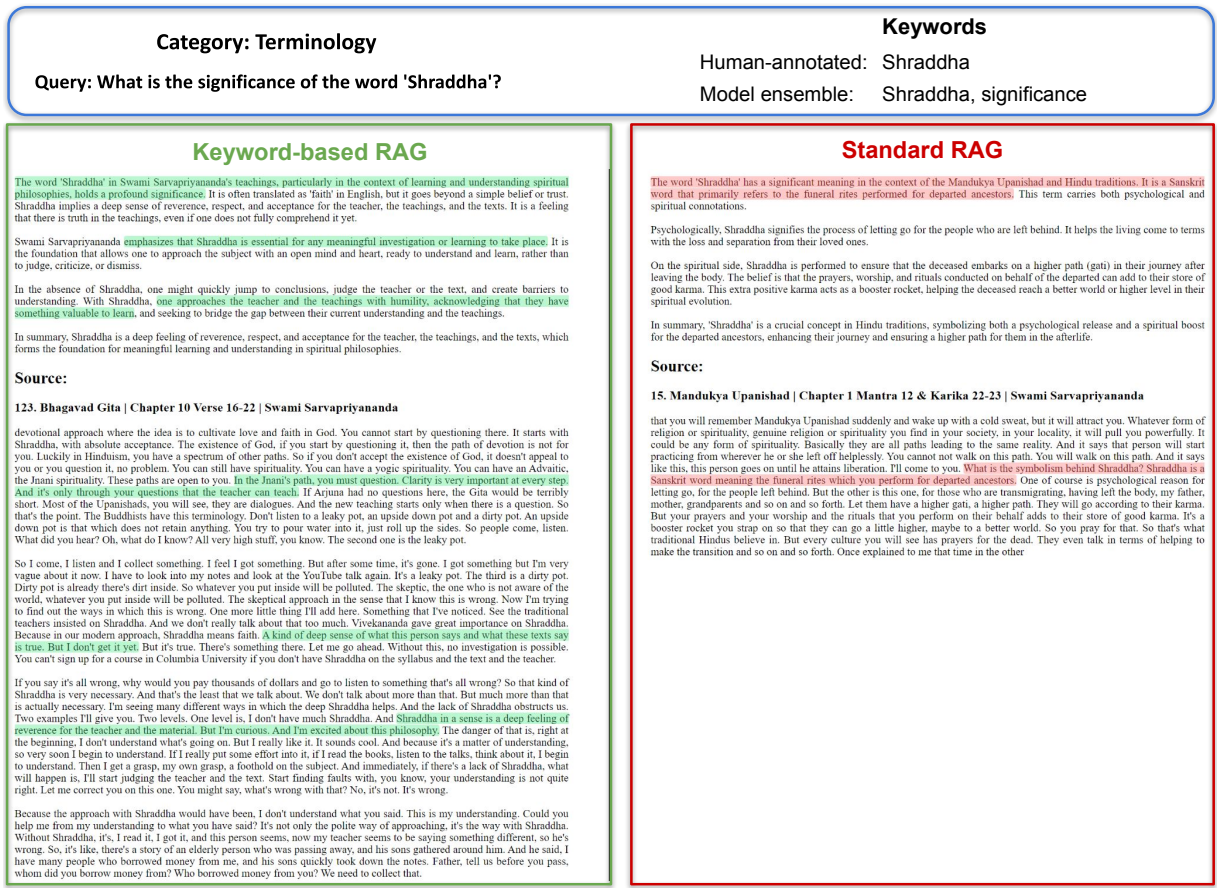

Figure 13: **Keyword-RAG vs Standard-RAG: Terminology.** The keyword-RAG model retrieves a comprehensive exposition on the concept of Shraddha, loosely translated as conviction, in the context of qualifications for the study of Advaita Vedanta. The standard RAG although retrieves a passage containing the word, it is however not directly related to what the questioner intends. This seems to be an unfortunate case of false positive for standard RAG due to inadequate or implied meaning in the query.

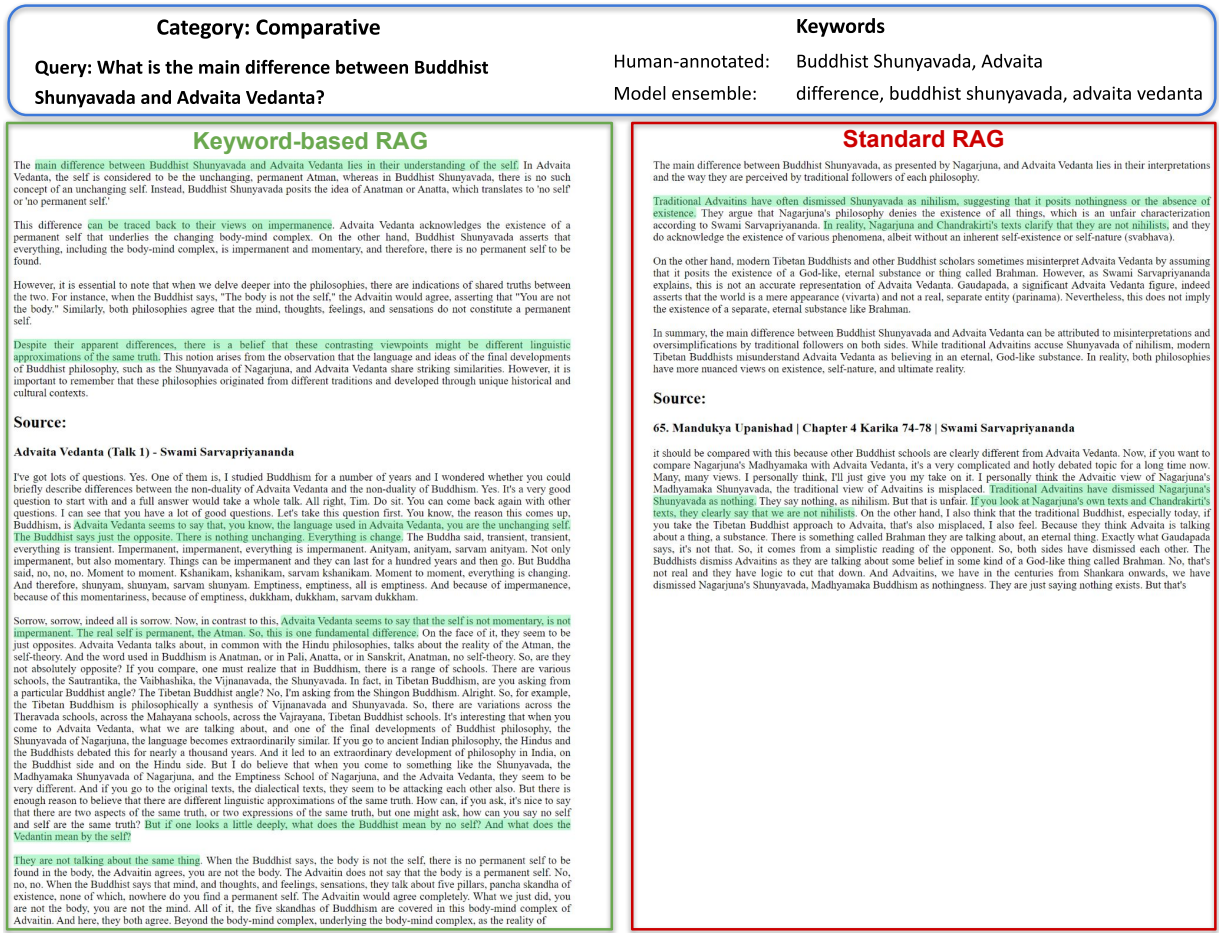

Figure 14: **Keyword-RAG vs Standard-RAG: Comparative.** This is a case where both models retrieve meaningful passages, with the keyword-based model providing greater context due to retrieval expansion. The standard model also seems to be using outside knowledge not mentioned in the passage.

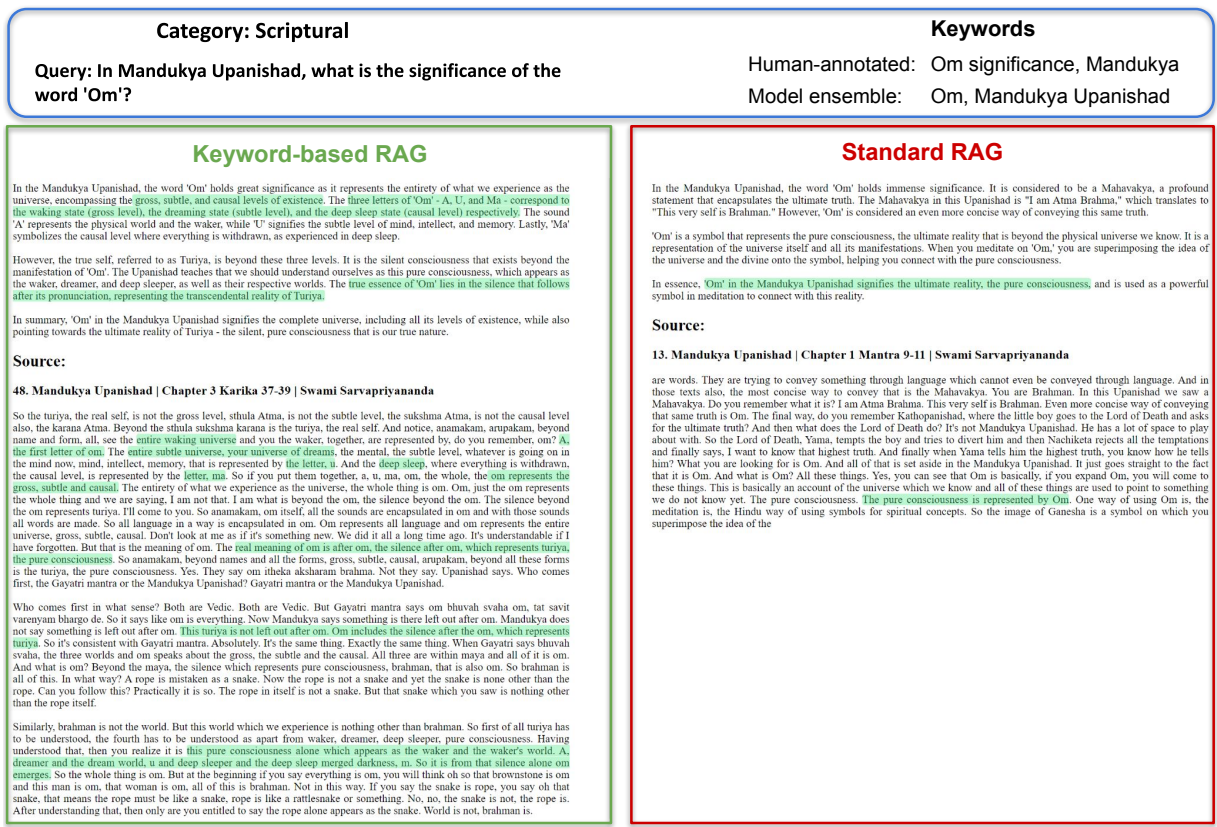

Figure 15: **Keyword-RAG vs Standard-RAG: Scriptural.** Both models retrieve from the correct scripture. However, while the retrieved passage in keyword-RAG is very comprehensive and accurately captures the context of the question, the standard RAG misses the point of the question and gives a generic answer.

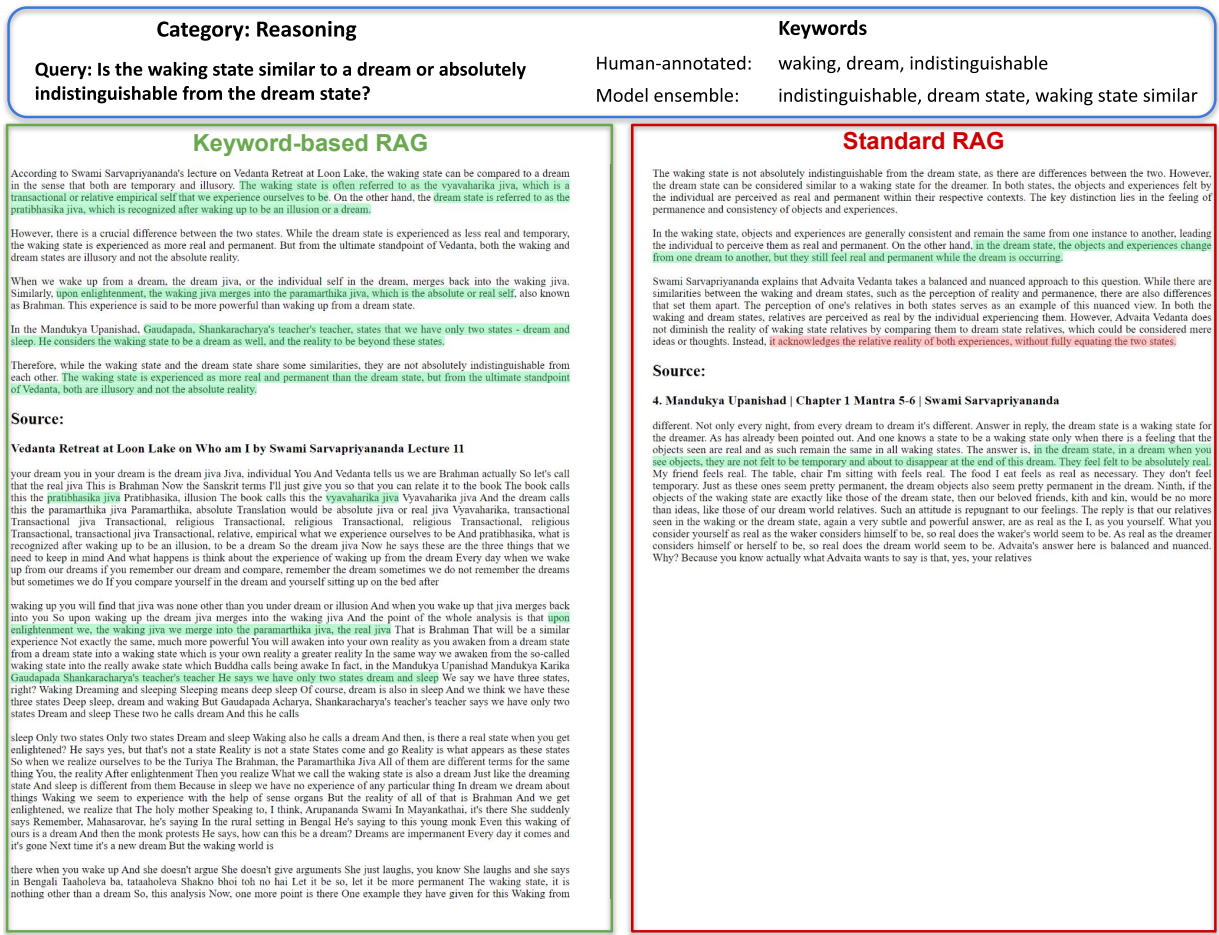

Figure 16: **Keyword-RAG vs Standard-RAG: Reasoning.** The retrieved passage in keyword-RAG is technical and comprehensive and the generated answer effectively summarizes the main points. The standard model is also good, although the explanation is not as effective owing to the quality of retrieval.